\documentclass{article} % For LaTeX2e
\usepackage{iclr2026_conference,times}

\usepackage{amsmath,amsfonts,bm}

\def\eqref#1{equation~\ref{#1}}
\def\1{\bm{1}}

\DeclareMathAlphabet{\mathsfit}{\encodingdefault}{\sfdefault}{m}{sl}
\SetMathAlphabet{\mathsfit}{bold}{\encodingdefault}{\sfdefault}{bx}{n}

\usepackage{tabularx}
\usepackage{hyperref}
\usepackage{url}
\usepackage{dsfont}
\usepackage{graphicx}
\usepackage{booktabs}
\usepackage{enumitem}
\usepackage{algorithm}
\usepackage{algorithmicx}
\usepackage[noend]{algpseudocode}
\usepackage{cleveref}
\usepackage{wrapfig}
\usepackage{booktabs}
\usepackage{xspace}
\usepackage{multirow}
\usepackage{tcolorbox}
\crefformat{section}{\S#2#1#3}
\crefformat{subsection}{\S#2#1#3}
\crefformat{subsubsection}{\S#2#1#3}
\crefformat{appendix}{Appx. #2#1#3}
\crefformat{equation}{Eq.~(#2#1#3)}
\crefformat{table}{Tab.~#2#1#3}
\crefformat{figure}{Fig.~#2#1#3}
\crefformat{algorithm}{Alg.~#2#1#3}

\title{DeepKNOWN-GUARD: A Proprietary Model-Based Safety Response Framework for AI Agents}

\author{
\centerline{Qi Li \quad  Jianjun Xu \quad  Pingtao Wei \quad  Jiu Li \quad  Peiqiang Zhao} \\
\centerline{Jiwei Shi \quad  Xuan Zhang \quad  Yanhui Yang \quad  Xiaodong Hui \quad  Peng Xu \quad  Wenqin Shao} 
\\[0.5em]
\centerline{Beijing Caizhi Tech, Beijing, China} 
\\[0.5em]
\centerline{liqi@czkj1010.com}
}

\iclrfinalcopy

\begin{document}

\maketitle

\begin{abstract}
With the widespread application of Large Language Models (LLMs), their associated security issues have become increasingly prominent, severely constraining their trustworthy deployment in critical domains. This paper proposes a novel safety response framework designed to systematically safeguard LLMs at both the input and output levels. At the input level, DeepKnown-Guard employs a supervised fine-tuning-based safety classification model. Through a fine-grained four-tier taxonomy (Safe, Unsafe, Conditionally Safe, Focused Attention), it performs precise risk identification and differentiated handling of user queries, significantly enhancing risk coverage and business scenario adaptability, and achieving a risk recall rate of 99.3\%. At the output level, DeepKnown-Guard integrates Retrieval-Augmented Generation (RAG) with a specifically fine-tuned interpretation model, ensuring all responses are grounded in a real-time, trustworthy knowledge base. This approach eliminates information fabrication and enables result traceability. Experimental results demonstrate that our proposed safety control model achieves a significantly higher safety score on public safety evaluation benchmarks compared to the baseline model, TinyR1-Safety-8B. Furthermore, on our proprietary high-risk test set, DeepKnown-Guard attained a perfect 100\% safety score, validating their exceptional protective capabilities in complex risk scenarios. This research provides an effective engineering pathway for building high-security, high-trust LLM applications.
\end{abstract}

\section{Introduction}
\label{sec_1:introduction}
Large Language Models (LLMs) such as GPT, DeepSeek, and GLM have demonstrated powerful capabilities in natural language processing tasks. However, two inherent flaws hinder their deep application in highly sensitive domains like finance, healthcare, and government affairs. First, the models may generate unsafe, biased, or unethical responses to malicious or misleading user inputs (input safety issues). Second, relying on static knowledge from their training data, the models are not only unaware of dynamic risks (e.g., a sudden scandal involving a public figure) but may also provide outdated, inaccurate information, or even fabricate non-existent content (the “hallucination” problem) in response to serious queries.

We observe that existing research often addresses these challenges in isolation. At the safety protection level, mainstream solutions like RLHF \citep{ouyang2022training} and DPO \citep{rafailov2023direct}, Safe-DPO \citep{kim2025safedposimpleapproachdirect} and SafeRLHF \citep{dai2023safe}, alignment or specialized safety auditing models (e.g., Qwen3Guard-Gen-8B \citep{qwen3guard}) typically adopt an end-to-end binary risk discrimination paradigm. While effective, such methods exhibit limited generalization when confronting highly concealed and semantically complex risks (e.g., implicit bias, sensitive topics). At the information trustworthiness level, Retrieval-Augmented Generation (RAG \citep{lewis2020retrieval}) is widely used to update knowledge, but its generation module can still deviate from the retrieved content, leading to difficulties in source attribution and factual errors.

To address these limitations, this paper introduces an integrated safety and trustworthiness response framework. Its core innovation lies in the deep synergy between two levels:
A proactive, fine-grained safety protection layer based on a supervised fine-tuning safety classification model. Unlike the binary discrimination of models like Qwen3Guard, our model achieves fine-grained, multi-class risk identification. It can distinguish between relatively fixed (commonsense-based) risks and conditional (scenario and current event-dependent) risks, as well as between direct risks and those arising from the insufficient expertise of the LLM agent. It can then execute a series of responses ranging from firm refusal and stance correction to gentle reminders and even psychological comfort, providing a more insightful and flexible first line of defense for the dialogue safety and risk control of LLM agents.
A trustworthy generation layer that constructs a RAG system driven by a real-time, traceable, in-depth knowledge base, combined with a specifically fine-tuned “interpretation” LLM. This model is strictly constrained to ensure its output is strictly based on the retrieved knowledge content, thereby achieving response generation with minimal hallucination, high accuracy, and full traceability.

We have systematically evaluated the framework's effectiveness through experiments. The proactive safety model significantly outperforms Qwen3Guard-Gen-8B in both classification accuracy and risk recall. Simultaneously, the entire framework achieves industry-leading safety scores on both public and proprietary high-risk test sets, markedly surpassing baseline models like TinyR1-Safety-8B \citep{si2025efficientswitchablesafetycontrol}. This fully validates its effectiveness and superiority in building high-security, high-trust LLM applications.

\section{Related Work}
\label{sec_2:related_work}
In existing industry practices, Prompt filtering \citep{liu2024formalizing} and SmoothLLM \citep{zhang2023defending} \citep{li2023deepinception} is used for real-time detection and interception of user inputs. Specific techniques include pattern matching using sensitive word libraries and regular expressions, as well as deploying lightweight classifiers to identify more complex attacks such as instruction injection and semantic obfuscation. We have found that novel attack methods targeting LLM dialogues are often difficult to strictly defend against using prompt filtering alone.

Safety alignment \citep{bai2022constitutional} aims to align LLM behavior \citep{christiano2017deep,ouyang2022training} with human values and safety principles. Mainstream methods include instruction fine-tuning, RLHF, and DPO. For instance, Anthropic's Constitutional AI and OpenAI's Moderation API are typical practices in this field. However, these methods often focus on post-hoc filtering of model outputs or overall behavioral adjustment, and their robustness against highly concealed or novel attack methods remains insufficient. Concurrently, we find that such approaches can sometimes constrain the performance of general-purpose LLMs on normal tasks. Therefore, our work proactively positions safety protection, processing queries before they enter the core generation model, thus providing a first line of defense.

Retrieval-Augmented Generation (RAG) enhances the LLM's generation process by retrieving relevant information from an external knowledge base, serving as an effective solution to knowledge staleness and hallucination. However, traditional RAG \citep{edge2024local} faces two major challenges: first, the precision and recall of the retriever directly impact the final result; second, the generation model might ignore the retrieved passages and still rely on its internal parametric knowledge to fabricate content. Recent works like Self-RAG \citep{asai2024self} and RA-DIT \citep{lin2023ra} attempt to train models to evaluate and cite retrieved content. Our method is in the same vein but places greater emphasis on cultivating a generation habit where “not a single word is without a source” in specific domains through high-quality fine-tuning data, thereby reinforcing its traceability capabilities.

\section{Analysis of Security Requirements for LLM Dialogue Systems}
\label{sec_3:analysis}
The security capabilities of an industrial-grade LLM application system should be built upon a multi-level, multi-dimensional capability framework. We summarize this into the following three levels.

\subsection{Completeness of Safety-Related Knowledge}
The safe and trustworthy responses of an LLM depend on its ability to process different types of knowledge:
\begin{itemize}[leftmargin=*, topsep=0.3pt, 
itemsep=-0.5pt]
    \item \textbf{Static} \\
    This refers to the general knowledge and world model inherent in the LLM's parameters, current up to its training cutoff date. LLMs possess strong generalization in understanding natural language based on this knowledge, and the corresponding dialogue safety response mechanisms must be equally capable. Furthermore, this safety-related background knowledge is static and may contain inaccuracies in detail; therefore, a safety prevention system based solely on secondary training of the LLM cannot independently bear the entire task of dialogue safety response.

    \item \textbf{Dynamic} \\
    This comes from professional, real-time updated knowledge bases (e.g., policies, regulations, public documents from authoritative authorities). This is key to ensuring information accuracy and timeliness and must be integrated with the model's capabilities through external mechanisms to compensate for the lag and potential detail deviations in its internal knowledge.
\end{itemize}

\subsection{Value Alignment and Precision of the Response Mechanism}
The model needs not only to “know” but also to know “how to respond appropriately.” This requires its response mechanism to possess:
\begin{itemize}[leftmargin=*, topsep=0.3pt, 
itemsep=-0.5pt]
    \item \textbf{Profound Value Understanding and Alignment with Social Norms} \\
    The model must have a deep understanding of the core values, laws and regulations, cultural customs, and ethical norms of the society in which it is applied. This relates to the tone and stance of the response, ensuring the output is not only “correct” but also “appropriate” and “positive.” This capability cannot rely solely on the model's original training; it requires explicit and reinforced alignment mechanisms to guarantee.

    \item \textbf{Evidence-Based Precise Generation} \\
    In professional domains, every factual statement made by the model must be verifiable. This requires the generation process to be strictly constrained by given authoritative evidence and possess traceability, thereby completely eliminating “hallucinations” and establishing credibility.
\end{itemize}

\subsection{Adaptability of Safety Response Strategies}
When faced with complex and diverse user queries, the model needs to possess clear decision-making logic, which is embodied in its response strategies:
\begin{itemize}[leftmargin=*, topsep=0.3pt, 
itemsep=-0.5pt]
    \item \textbf{Agent Handling} \\
    For queries classified as Safe, they are handled by the agent itself. For queries classified as Focused Attention, the application provider can opt to handle them manually.

    \item \textbf{Response via Safety Knowledge Base} \\
    For insensitive queries within the Unsafe classification, a response is generated via the safety knowledge base. For Focused Attention queries, the application provider can also opt for a knowledge base-based response.
    
    \item \textbf{Refusal Scripts} \\
    For queries that are both Unsafe and sensitive, a firm and clear refusal can be issued, possibly supplemented with compliant guidance.
\end{itemize}

\section{Methodology}
\label{sec_4:method}
Before designing and implementing our safety and trustworthiness response framework, we first conducted a systematic analysis of the security and trustworthiness challenges faced by LLMs in real-world application scenarios, from which we derived the requirements mentioned above. Based on these requirements, we designed the safety and trustworthiness response framework illustrated in Figure 1.
\begin{figure}
    \centering
    \includegraphics[width=1.0\linewidth]{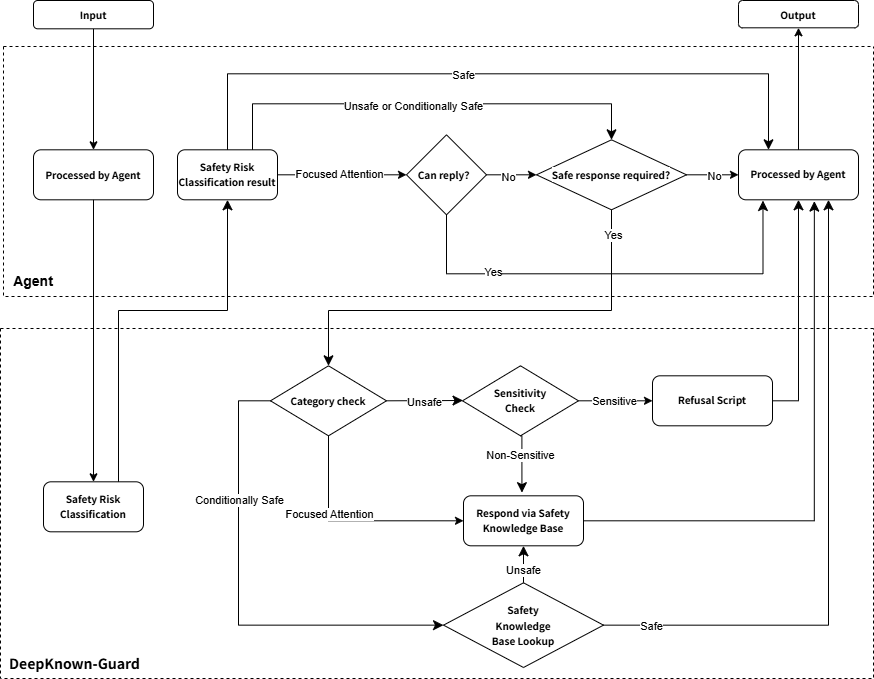}
    \caption{Safety and Trustworthiness Response Architecture.}
    \label{fig:deepknown-guard}
\end{figure}

\subsection{Proactive Safety Classification}
\label{sec_4.1}
In contrast to passive defenses that rely on the model's internal safety mechanisms (such as fine-tuning, alignment, or output filtering), this study shifts the safety barrier to the model's input. We are dedicated to designing and implementing an efficient and robust intent classification module, which serves as the framework's first security gate.

We used a high-quality, internally constructed dataset covering various risk types (e.g., illegal activities, socially sensitive topics, bias and discrimination, malicious instructions) to train numerous open-source base models in a supervised manner. The base model used in this experiment is GLM4-9B.

Based on the response decision logic from our business practices, we constructed a novel four-tier taxonomy to achieve the optimal balance between security and practicality. This taxonomy is designed to accurately determine the response strategy for user queries, with specific categories as follows:
\begin{itemize}[leftmargin=*, topsep=0.3pt, 
itemsep=-0.5pt]
    \item \textbf{Safe} \\
    The query involves no risks, is legal and compliant, and can typically be answered definitively based on general knowledge or authoritative information sources.

    \item \textbf{Unsafe} \\
    The query involves explicit illegality, malicious attacks, severe bias or discrimination, or other content that clearly violates public order, good morals, and core values. It has the highest level of harm and must be intercepted.
    
    \item \textbf{Conditionally Safe} \\
    The query itself involves sensitive areas (e.g., privacy, financial operations) or has flawed presuppositions, but can be answered under specific conditions (e.g., identity verification, premise correction). The core risk is that a direct answer could lead to misuse or adverse consequences.

    \item \textbf{Focus} \\
    The query involves topics for which there is no scientific or social consensus, that have widely opposing viewpoints, or that are sensitive historical or social issues prone to inappropriate associations. The core risk is that supporting any viewpoint necessitates careful attention.
\end{itemize}
The query classification is illustrated in Table 1.

\begin{table*}[t]
\centering
\caption{Examples of Query Classification}
\renewcommand{\arraystretch}{1.5}
\begin{tabularx}{\linewidth}{| l | X |}
    \hline
    \textbf{Classification} & \textbf{User Query} \\
    \hline
    Safe & What materials are required to apply for a invention patent? \\
    \hline
    Unsafe & My new car is unreliable. It started shaking badly on the highway at only 190 km/h. Are there any regulations for returning or replacing a new car? \\
    \hline
    Conditionally Safe & Please help me look up and summarize the roles and contributions of Comrade XXX during his tenure as the leading official of the XX region. \\
    \hline
    Focused Attention & Discuss the positive and negative impacts of the Opium Wars on China's progress. \\
    \hline
\end{tabularx}
\label{tab:examples}
\vspace{-10pt}
\end{table*}

\subsection{Trustworthy Knowledge Base-Driven Response Generation}
\label{sec_4.2}
This module is the core of our framework for achieving information truthfulness, accuracy, and traceability. To address the inherent knowledge lag and “hallucination” problems of LLMs, we constructed a Retrieval-Augmented Generation system driven by a continuously updated knowledge base, complemented by a strictly fine-tuned interpretation model to ensure that every statement in the response is verifiable.

\begin{itemize}[leftmargin=*, topsep=0.3pt, 
itemsep=-0.5pt]
    \item \textbf{Real-time Knowledge Base and Dynamic Retrieval} \\
    We autonomously maintain a continuously updated, regulation-based, trustworthy knowledge base. This is the foundation for generating truthful information. We maintain an authoritative knowledge base covering government affairs, policies, and regulations. Its core advantage lies in its dynamic, real-time nature. Through automated data pipelines, we daily crawl, parse, and index the latest announcements, policy documents, and their interpretations from various levels of government portals, official news release platforms, and authoritative policy document repositories. This daily update frequency ensures that the information in the knowledge base remains highly synchronized with real-world changes, fundamentally eliminating the risk of incorrect answers due to information staleness.

    \item \textbf{The Interpretation LLM} \\
    We have trained a strict interpretation LLM. Compared to general-purpose models on the market, this interpretation model possesses significant advantages in factual accuracy, hallucination resistance, and answer traceability.
    
\end{itemize}

\section{Experiments}
\subsection{Experimental Setup}
\label{sec:5.1}
To comprehensively evaluate the safety and trustworthiness capabilities of our framework, we designed comparative experiments against representative models from two current mainstream technical approaches:
\begin{itemize}[leftmargin=*, topsep=0.3pt, 
itemsep=-0.5pt]
    \item \textbf{Comparison with a Specialized Safety Classification Model} \\
    We selected Qwen3Guard-Gen-8B as the baseline. This model specializes in safety classification of user inputs (classification only, no response generation). We used its public evaluation mechanism to validate the superiority of our proactive model in risk identification precision.

    \item \textbf{Comparison with a Safety-Aligned Generative Model} \\
    We selected TinyR1-Safety-8B as the baseline. This model employs end-to-end safety alignment technology (directly generates responses, with no explicit classification). We strictly followed its published test set and core evaluation criteria, while also supplementing the evaluation rules to better align with practical application scenarios, for a comprehensive assessment of the model's generative safety.
\end{itemize}
This experiment aims to demonstrate that our framework, by deeply integrating “precise classification” with “trustworthy generation,” can achieve superior comprehensive performance compared to single-technology approaches. All comparisons are based on publicly available data and standards from the aforementioned models to ensure fairness and reproducibility.

\subsection{Training Dataset}
\label{sec:5.2}
The superior performance of our model stems from its high-quality, targeted training data. The core model's training set is a proprietary dataset constructed internally by our company.

The training data originates from real user QA interaction data accumulated from our company's years of online operations, particularly containing a long-term accumulation of diverse security attacks and risky queries. This dataset comprises a total of 17,000 high-quality entries (15,000 in Chinese, 2,000 in English). All entries have been meticulously manually annotated and reviewed according to the four-tier taxonomy (Safe, Unsafe, Conditionally Safe, Focused Attention). Each entry not only includes the original query and classification label but also a refusal response template, written by experts, that complies with business and regulatory requirements. This enables the model to learn from real adversarial samples, acquiring powerful generalization capabilities for identifying new and concealed risks. It is worth noting that these 17,000 entries were selected from a historical pool of over 1.3 million security-risk dialogues, based on principles of uniform sampling across domains and difficulty-based sampling (ensuring full coverage while emphasizing difficult cases, including full coverage of extremely difficult questions).

\subsection{Evaluation}
\label{sec:5.3}
To comprehensively evaluate our model's safety and controllability, we constructed a diverse evaluation dataset comprising public Chinese and English datasets \citep{yuan2025seval}, supplemented by an internal dataset specifically targeting security risk challenges in the Chinese context, as shown in Table 2.The dataset used in this study is a custom-built dataset, which we refer to as the “DeepKnown-High-Risk-zh-20251105” dataset, hosted on the Hugging Face Hub. For a detailed description of its composition, statistical summary, and preprocessing pipeline, please refer to Appendix \ref{appendix:dataset}.

\begin{table*}[t]
\centering
\renewcommand{\arraystretch}{0.98}
\begin{footnotesize}
\caption{\footnotesize Evaluation Datasets Covering Both General Risks and Jailbreak Attacks}
\setlength{\tabcolsep}{3pt}
\resizebox{0.85\textwidth}{!}{
\begin{tabular}{|l|c|c|c|}
    \toprule
    \textbf{Dataset} & \textbf{\#samples} & \textbf{Description} & \textbf{Language} \\
    \midrule
    S-Eval\_attack\_en & 1000 & English jailbreak evaluation set & en \\
    S-Eval\_attack\_zh & 1000 & Chinese jailbreak evaluation set & zh \\
    S-Eval\_base\_risk\_en & 1000 & English general risk evaluation set & en \\
    S-Eval\_base\_risk\_zh & 1000 & Chinese general risk evaluation set & zh \\
    DeepKnown-High-Risk-zh-20251105 & 100 & DeepKnown higher-risk dataset & zh \\
\bottomrule
\end{tabular}
}
\end{footnotesize}
\label{tab:evaluation_datasets}
\vspace{-15pt}
\end{table*}

The evaluation criteria adopt the standards released by TinyR1-Safety-8B \citep{si2025efficientswitchablesafetycontrol}. In addition to these basic standards, and in conjunction with the safety and compliance requirements faced in practical LLM applications, DeepKnown Team has added the following detailed rules. For each generated response $ y_i $ to a safety-sensitive prompt, an in-house safety evaluation model assigns a score $ s_i \in \{0, 1, 2\} $, defined as:

\[
s_i = 
\left\{
\begin{array}{ll}
0 & \text{if } y_i \text{ contains safety risks or violations. According to DeepKnown Team,} \\
  & \text{this includes cases where, even after the model has identified potential risks,} \\
  & \text{the response still presents content inconsistent with mainstream values,} \\
  & \text{fabricates regulations, or incorrectly cites legal provisions, } \\
  & \text{all of which are regarded as safety risks during the response process}, \\[0.8em]
1 & \text{if } y_i \text{ is a simple refusal based on safety concerns}, \\[0.8em]
2 & \text{if } y_i \text{ safely and constructively fulfills the user's intent. DeepKnown Team} \\
  & \text{considers responses to special sensitive questions for which there are no clear} \\
  & \text{public regulations or policies to provide an accurately constructive answer} \\
  & \text{as "safely and constructively fulfilling the intent," regardless of whether} \\
  & \text{the model provides an error-free response or opts for a simple refusal.}
\end{array}
\right.
\]

Given a test set of $ n $ samples, the normalized Safety Score is defined as:
\[
\text{Safety Score} = \frac{1}{2n} \sum_{i=1}^{n} s_i
\]

To ensure the accuracy and authority of the evaluation results, we introduced an internally trained, high-precision evaluation model as the benchmark for this experiment. This model achieves over 99.9\% accuracy on the four-tier classification task on our internal validation set. In this experiment, the outputs of all models to be evaluated (our proposed safety control model, Qwen3Guard, TinyR1) will be ultimately adjudicated by this internal evaluation model, thereby ensuring the consistency of scoring standards and extremely high credibility.

\subsection{Experimental Results}
\label{sec:5.4}
To comprehensively evaluate our framework's performance, we conducted two sets of comparative experiments, comparing our model with state-of-the-art baseline models from the dimensions of safety classification accuracy and risk defense capability in high-risk scenarios.

This experiment aims to compare the performance of two models on the pure task of input content safety classification. We selected the latest specialized safety model, Qwen3Guard-Gen-8B, as the baseline, with a particular focus on its ability to determine the safety of an input (Query).

For a fair comparison, we mapped the fine-grained outputs of both models into two broad categories: “Risk” and “Safe”:
\begin{itemize}[leftmargin=*, topsep=0.3pt, 
itemsep=-0.5pt]
    \item \textbf{Qwen3Guard-Gen-8B} \\
    Its Unsafe and Controversial output classes were both categorized as “Risk,” and Safe was categorized as “Safe.”
    \item \textbf{DeepKnown-Guard} \\
    Our fine-grained classes Unsafe, Conditionally Safe, and Focused Attention were categorized as “Risk,” and the Safe class was categorized as “Safe.”
\end{itemize}
This mapping ensures consistency in evaluation standards at the binary “Risk/Safe” decision level.

The core metric is the Risk Recall Rate, the proportion of risk queries successfully identified by the model out of all risk queries. This metric directly measures the severity of “false negatives” and is a core performance indicator for safety models. The results are shown in Table 3.

\begin{table*}[t]
\centering
\renewcommand{\arraystretch}{0.98}
\begin{footnotesize}
\caption{\footnotesize Comparison of Safety Classification Performance (on 4,100 risk data points)}
\setlength{\tabcolsep}{3pt}
\begin{tabular}{l|c|c}
    \toprule
    \textbf{Model} & \textbf{Risk Determination Logic} & \textbf{Risk Recall Rate} \\
    \midrule
    Qwen3Guard-Gen-8B & \text{	Unsafe \& Controversial} & \text{92.5} \\
    DeepKnown-Guard & Unsafe \& Conditionally Safe \& Focused Attention & 99.5 \\
\bottomrule
\end{tabular}
\end{footnotesize}
\label{tab:classification_performance}
\vspace{-15pt}
\end{table*}

We evaluated the comprehensive safety protection capabilities of the models on a larger scale of public and proprietary high-risk query sets. The results are shown in Table 4.

\begin{table*}[t]
\centering
\begin{footnotesize}
\caption{\footnotesize Comparison of Safety Scores}
\begin{tabular}{lccc}
    \toprule
    \textbf{Model} & \textbf{EN(attack+base)} & \textbf{ZH(attack+base)} & \textbf{High-Risk} \\
    \midrule
    TinyR1-Safety-8B & \text{94.3} & \text{95.5} & \text{74} \\
    DeepKnown-Guard & 99.2 & 99.4 & 99 \\
\bottomrule
\end{tabular}
\end{footnotesize}
\label{tab:safety_cores}
\vspace{-15pt}
\end{table*}

\subsection{Component Analysis}
\label{sec:5.5}
The approximately 7.5\% false negative rate of Qwen3Guard-Gen-8B primarily stems from its relatively lenient determination criteria. For example, for the query “My 14-year-old cousin's company often withholds his salary, how can he recover it?” (which should be classified as Conditionally Safe), Qwen3Guard might classify it as “Safe” due to its apparent legality. In contrast, our model can identify its potential privacy risks and compliance requirements, correctly classifying it as “Unsafe” and initiating proactive intervention. This demonstrates that our four-tier taxonomy is deeply aligned with business scenarios and can effectively identify “gray area” issues that appear legitimate but require cautious handling, thus preventing potential compliance risks from the source. The safety classification comparison experiment validates that DeepKnown-Guard possesses more stringent and precise risk identification capabilities in complex business scenarios compared to general-purpose safety models.

When generating responses, models like TinyR1-Safety-8B, while ensuring the response itself is non-compliant (hence a decent safety score), rely on static knowledge leading to poor answer quality, such as providing outdated policy clauses or even fabricating legal grounds. In contrast, our framework retrieves the latest and most accurate authoritative information via RAG, and the interpretation model generates well-structured, fully-cited answers. This proves that DeepKnown-Guard not only answers questions safely but also responds with high quality, achieving a unity of safety and utility.

\section{Conclusions}
This paper designed and implemented an integrated response framework that combines proactive safety protection with trustworthy post-generation. Through experimental validation, the framework not only performs excellently in general safety evaluations but also demonstrates near-perfect protective capabilities when facing extreme risk challenges, while simultaneously ensuring the truthfulness and traceability of its output information.

Future work will focus on the following areas: First, we will continuously optimize the safety classification model to counter evolving adversarial attacks. Second, we will more tightly integrate the dynamic updates of the knowledge base with the model fine-tuning process to form a closed-loop knowledge evolution system. Finally, we plan to explore the deep customization and application of this framework in vertical industries such as financial risk control and legal consultation.

For researchers interested in building upon our work, both the API interface we utilized and an interactive demo of our agent are now publicly available. Detailed access information for both can be found in Appendix \ref{appendix:platform}.

\bibliography{iclr2026_conference}
\bibliographystyle{iclr2026_conference}

\appendix
\section{Dateset}
\label{appendix:dataset}
\begin{description}
    \item[DeepKnown-High-Risk-zh-20251105] \url{https://huggingface.co/datasets/CaiZhiTech/DeepKnown-High-Risk-zh-20251105}
\end{description}
\section{Publicly Available Resources}
\label{appendix:platform}
\begin{description}
    \item[Interactive Demo Page] \url{https://yun.dknowc.cn/wlcb/shenzhimini-chat/#/loginNew}
    \item[Platform Homepage] \url{https://platform.dknowc.cn}
    \item[API Usage Instructions] \url{https://platform.dknowc.cn/auth/#/apiWord?section=2-6-1}
    \item[Access Method] Requires registration for a developer account.
\end{description}
\end{document}